# Solving POMDPs by Searching the Space of Finite Policies

**Nicolas Meuleau, Kee-Eung Kim, Leslie Pack Kaelbling** and **Anthony R. Cassandra**
Computer Science Dept, Box 1910, Brown University, Providence, RI 02912-1210
{nm, kek, lpk, arc}@cs.brown.edu

## Abstract

Solving partially observable Markov decision processes (POMDPs) is highly intractable in general, at least in part because the optimal policy may be infinitely large. In this paper, we explore the problem of finding the optimal policy from a restricted set of policies, represented as finite state automata of a given size. This problem is also intractable, but we show that the complexity can be greatly reduced when the POMDP and/or policy are further constrained. We demonstrate good empirical results with a branch-and-bound method for finding globally optimal deterministic policies, and a gradient-ascent method for finding locally optimal stochastic policies.

## 1 INTRODUCTION

In many application domains, the partially observable Markov decision process (POMDP) [1, 22, 23, 5, 9, 4, 13] is a much more realistic model than its completely observable counterpart, the classic MDP [11, 20]. However, the complexity resulting from the lack of observability limits the application of POMDPs to dramatically small decision problems. One of the difficulties of the optimal—Bayesian—solution technique is that the policy it produces may use the complete previous history of the system to determine the next action to perform. Therefore, the optimal policy may be infinite and we have to approximate it at some level to be able to implement it in a finite machine. Another problem is that the calculation requires reformulating the problem in the continuous space of belief functions, and hence it is much harder than the simple finite computation that is sufficient to optimize completely observable MDPs.

What can we do if we have to solve a huge POMDP? Since it may be impossible just to represent the optimal policy in memory, it makes sense to restrict our search to policies that are reasonable in some way (calculable and storeable in a finite machine). Knowing that any policy is representable as a (possibly infinite) state automaton, the first constraint we would want to impose on the policy is to be representable by a *finite* state automaton, or, as we will call it, a finite "policy graph". Many previous approaches implicitly rely on a similar hypothesis: Some authors [14, 12, 2, 27] search for optimal reactive (or memoryless) policies, McCallum [15, 16] searches the space of policies using a finite-horizon memory, Wiering and Schmidhuber's HQL [26] learns finite sequences of reactive policies, and Peshkin *et al.* [19] look for optimal finite-external-memory policies. All these examples are particular cases of finite policy graphs, with a set of extra structural constraints in each case (i.e., *not* every node-transition and action choice is possible in the graph). Note that in general, finite policy graphs can remember events arbitrarily far in the past. They are just limited in the number of events they can memorize.

In this paper we study the problem of finding the best policy representable as a finite policy graph of a given size, possibly with simple constraints on the structure of the graph. The idea of searching explicitly for an optimal finite policy graph for a given POMDP is not new. In the early 70s, Satia and Lave [21] proposed a heuristic approach for finding ε-optimal decision trees. Hansen [6, 7, 8] proposed a policy iteration algorithm that outputs an ε-optimal controller. These solution techniques work explicitly in the belief space used in the classical—Bayesian—optimal solution of POMDPs, and they output policy-graphs which are not more than ε *from this optimal solution.* Another approach uses EM to find controllers that are optimal over a finite horizon [10].

A characteristic property of our algorithms is that they scale up well with respect to the size of the problem. Their drawback is that their execution time increases quickly with the size of the policy graph, i.e., with the complexity of the policy we are looking for. In general, they will be adapted to large POMDPs where relatively simple policies perform near optimally. Another characteristic of our approach is that we do not refer to the value of the optimal Bayesian solution anymore, we just want the best graph given the constraint imposed on the number of nodes. Note that the optimality criterion used is the same as in the Bayesian approach, i.e., the expected discounted cumulative rewards (the expectation being relative to the prior belief on the states). However, since we do not evaluate the solution produced *relative to the optimal performance*, the problem may be solved without using the notion of belief-space.

Our development relies on a basic property of finite-state controllers that has already been stressed by Hansen [6], and that is also very close to Parr and Russell's HAM theorem [18]. Namely, given a POMDP and a finite policy graph, the sequence of pairs (state of the POMDP, node of the policy graph) constitutes a Markov chain. Going farther, we define a new MDP on the cross product of the state-space and the set of nodes, where a decision is the choice of an action *and* of a next node (conditioned on the last observation). Working in this cross-product MDP presents many advantages: it allows us to calculate and then differentiate the value of a fixed policy, to calculate upper and lower bounds on the value attainable by completing a given partial policy, and also to establish some complexity results. We use these properties to develop implementations of two classic search procedures (one global and one local) where the majority of the computation consists of solving some Bellman equations in the cross-product MDP. An important point is that the structure of both the POMDP and the policy graph (if there is one) is reflected in the cross-product MDP. It can be used to accelerate the solution of Bellman equations [3], and hence the execution of the solution techniques. In other words, the algorithms we propose can exploit the structure of the POMDP to find relatively quickly the best general finite policy graph of a given size. If this leverage is not sufficient, we may limit further the search space by imposing some structure on the policy graph, and then using this structure to speed up the solution of the cross-product MDP (for instance, we can limit ourselves to one of the finite-memory architectures mentioned above).

The paper is organized as follows. First we give a quick introduction to POMDPs and policy graphs, and define the cross-product MDP. Then we show that finding the best deterministic finite policy graph is an NP-hard problem. There is then no really easy way to solve our problem. In this paper, we propose two possible approaches: a global branch and bound search for finding the best deterministic policy graph, and a local gradient descent search for finding the best stochastic policy graph. These two algorithms are based on solving some Bellman equations in the cross-product MDP. Therefore, they can take full advantage of any preexisting structure in the POMDP or in the policy graph. Typically, these algorithms will be adapted to very structured POMDPs with a large number of states, a small number of observations, and such that some simple policies perform well. In the end of the paper, we give empirical evidence that our approach allows the solution of some POMDPs whose size is far beyond the limits of classical solutions.

## 2    POMDPS AND FINITE POLICY GRAPHS

### 2.1    POMDPS

A partially observable Markov decision process (POMDP) is defined as a tuple $(S, O, A, B, T, R)$ where:

- $S$ is the (finite) set of states;

- $O$ is the (finite) set of observations;

- $A$ is the (finite) set of actions;

- $B(s, o) = \Pr(o^t = o \mid s^t = s)$ for all $t$;

- $T(s, a, s') = \Pr(s^{t+1} = s' \mid s^t = s, a^t = a)$ for all $t$;

- $r^t = R(s, a, s')$ if $s^t = s$, $a^t = a$ and $s^{t+1} = s'$, for all $t$.

The underlying Markov decision process (MDP) $(S, A, T, R)$ is optimized is the following way [11, 20]: given an initial state $s^0$, the aim is to maximize the expected discounted cumulative reward

$$\mathrm{E}\left(\sum_{t=0}^{\infty} \gamma^t r^t \mid s^0\right),$$

where $\gamma \in [0, 1)$ is the discount factor. The optimal solution is a mapping $\mu^* : S \to A$ that specifies the action to perform in each possible state. The optimal expected discounted reward, or "value function", is defined as the unique solution of the set of Bellman equations:

$$V^*(s) = \max_{a \in A}\left[\sum_{s' \in S} T(s, a, s')\left(R(s, a, s') + \gamma V^*(s')\right)\right],$$

for all $s$. It is a remarkable property of MDPs that there exists an optimal policy that always executes the same action in the same state. Unfortunately, this policy cannot be used in the partially observable framework, because of the residual uncertainty on the current state of the process.

In the POMDP framework, a policy is in general a rule specifying the action to perform at each time step as a function of the whole previous history, i.e., the complete sequence of observation-action pairs since time 0. A particular kind of policy, the so-called reactive policies (RPs), condition the choice of the next action only on the previous observation. Thus, they can be represented as mappings $\mu : O \to A$. Given a probability distribution $\pi^0$ over the starting state,

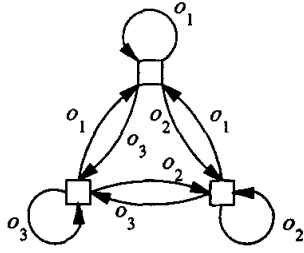

Figure 1: Structure of the policy graphs representing reactive policies ($|O| = 3$). The only degrees of freedom are the choices of the action in the $|O|$ nodes.

each policy $\mu$ (reactive or not) realizes an expected cumulated reward:

$$E\left(\sum_{t=0}^{\infty} \gamma^t r^t \mid \pi^0, \mu\right). \qquad (1)$$

The classical—Bayesian—approach allows us to determine the policy that maximizes this value. It is based on updating the state distribution (or belief) at each time step, depending on the most recent observations [5, 9, 4, 13]. The problem is re-formulated as a new MDP using belief-states instead of the original states. Generally, the optimal solution is not a reactive policy. It is a sophisticated behavior, with optimal balance between exploration and exploitation. Unfortunately, the Bayesian calculation is highly intractable as it searches into the continuous space of beliefs and considers every possible sequence of observations.

## 2.2 FINITE POLICY GRAPHS

A policy graph for a given POMDP is a graph where the nodes are labeled with actions $a \in A$, the arcs are labeled with observations $o \in O$, and there is one and only one arc emanating from each node for each possible observation. When the system is in a certain node, it executes the action associated with this node. This implies a state transition in the POMDP and eventually a new observation (which depends on the arrival state of the underlying MDP). This observation itself conditions a transition in the policy graph to the destination node of the arc associated with the new observation. Every policy has a representation as a possibly (countably) infinite policy graph. A policy that chooses a different action for each possible previous history will be represented by an infinite tree with a branch for each possible history. Reactive policies correspond to a special kind of finite policy graph with as many nodes as there are observations in the POMDP, and where all arcs labeled with the same observation go to the same node (figure 1).

In a stochastic policy graph there is a probability distribution over actions attached to each node instead of a single action, and transitions from one node to another are random, the arrival node depending only on the starting node

and the last observation. We will use the following notation:

- $N$ is the set of nodes of the graphs,

- $n^t \in N$ is the current node at time $t$,

- $\psi(n, a)$ is the probability of choosing action $a$ in node $n \in N$:

$$\psi(n, a) \stackrel{\text{def}}{=} \Pr(a^t = a \mid n^t = n), \text{ for all } t,$$

- $\eta(n, o, n')$ is the probability of moving from node $n \in N$ to node $n' \in N$, after observation $o \in O$:

$$\eta(n, o, n') \stackrel{\text{def}}{=} \Pr(n^{t+1} = n' \mid n^t = n \wedge o^{t+1} = o), \text{ for all } t.$$

- $\eta^0$ is the probability distribution of the initial node $n^0$ conditioned on the first observation $o^0$:

$$\eta^0(o, n) \stackrel{\text{def}}{=} \Pr(n^0 = n \mid o^0 = o).$$

In some cases, we will want to impose extra constraints on the policy graph. In most of this paper, we will limit ourselves to "restriction constraints" which consist in restricting the set of possible actions executable in some nodes, and/or restricting the set of possible successors of some nodes under some observations. Note that forcing the graph to implement an RP represents a set of restriction constraint as defined here. We consider more sophisticated sets of constraints in section 4.

## 2.3 THE CROSS-PRODUCT MDP

One advantage of representing the policy as a policy graph is that the cross-product of the POMDP and the policy graph is itself a finite MDP. Another interesting point is that all the structure of both the POMDP and the policy graph (if there is some) is represented in this cross-product MDP. It will allow us to develop relatively fast implementations of some classical techniques to solve our problem.

**Calculating the value of a policy graph**  The following theorem has been used by Hansen [6, 8, 7], and his closely related to Parr and Russell's HAM theorem [18].

**Theorem 1** *Given a policy graph $\mu = (\psi, \eta)$ and a POMDP $(S, O, A, B, T, R)$, the sequence of node-state pairs $(n^t, s^t)$ generated constitutes a Markov chain.*

The influence diagram of figure 2 proves this property: $(n^{t+1}, s^{t+1})$ depends only on $(n^t, s^t)$. The transition matrix $\tilde{T}^\mu$ of this Markov chain is given by

$$\tilde{T}^\mu((n, s), (n', s'))$$
$$= \sum_{a \in A} \psi(n, a) T(s, a, s') \sum_{o \in O} B(s', o) \eta(n, o, n'). \quad (2)$$

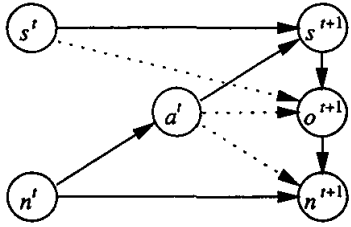

Figure 2: Influence diagram proving the Markov property of the cross-product MDP. Dotted arrows represent dependencies that we did not take into account in this work, but that are sometimes represented in other formulations. As shown, theorem 1 is still valid in this more general framework.

In the same way, we can calculate the expected immediate reward $\bar{C}^\mu$ associated with each pair $(n, s)$:

$$\bar{C}^\mu(n, s) = \sum_{a \in A} \psi(n, a) \sum_{s' \in S} T(s, a, s') R(s, a, s'). \quad (3)$$

Then the value function of the policy $\mu$ is found by solving the fundamental equation (in matrix form):

$$\bar{V}^\mu = \bar{C}^\mu + \gamma \bar{T}^\mu \bar{V}^\mu. \quad (4)$$

Since $\bar{T}^\mu$ is a stochastic matrix and $\gamma < 1$, the matrix $(I - \gamma \bar{T}^\mu)$ is invertible and we have:

$$\bar{V}^\mu = (I - \gamma \bar{T}^\mu)^{-1} \bar{C}^\mu. \quad (5)$$

Finally, the value of the policy, independent of the starting node and state, is

$$E^\mu = \bar{\pi}^0 \bar{V}^\mu, \quad (6)$$

where $\bar{\pi}^0$ is the joint probability distribution on $n^0$ and $s^0$:

$$\bar{\pi}^0(n, s) = \pi^0(s) \sum_{o \in O} B(s, o) \eta^0(o, n).$$

Differentiating this value with respect to the parameters of the graph will enable us to climb its gradient.

**Solving the cross-product MDP.** A complete MDP is defined on $\bar{S} = N \times S$. In each pair $(n, s)$, we have to choose an action $a$, wait for the new observation, and then chose the next node. It is equivalent to choosing an action and a mapping $\eta^n : O \rightarrow N$ which determines the next node as a function of the next observation. Therefore, the action space of the cross-product MDP is $\bar{A} = A \times N^O$.

**Theorem 2** *The tuple* $(\bar{S}, \bar{A}, \bar{T}, \bar{R})$ *is a finite stationary Markov decision process, where* $\bar{S} = N \times S$, $\bar{A} = A \times N^O$,

$$\bar{T}((n, s), (a, \eta^n), (n', s)) = T(s, a, s') \sum_{\substack{o \in O \text{ s.t.} \\ \eta^n(o) = n'}} B(s', o)$$

*and* $\bar{R}((n, s), (a, \eta^n), (n', s')) = R(s, a, s')$.

The fundamental equation of the MDP is, in matrix form:

$$\bar{V}^* = \max_{a \in A} \max_{\eta^n \in N^O} \left[ \bar{C} + \gamma \bar{T} \bar{V}^* \right],$$

where the maximization is applied row by row. When we expand this equation, the maximization over $\eta^n \in N^O$ can be replaced by a maximization over $n' \in N$ and moved to the end of the equation:

$$\bar{V}^*(n, s) = \max_{a \in A} \Big[ \sum_{s' \in S} T(s, a, s') \Big( R(s, a, s') + \gamma \sum_{o \in O} B(s', o) \max_{n' \in N} \bar{V}^*(s', n') \Big) \Big]. \quad (7)$$

The expected optimal reward, independent of the starting state and node, is $E^* = \bar{\pi}^0 \bar{V}^*$. The stationary optimal policy of the cross-product MDP is a mapping $\bar{\mu}^* : \bar{S} \rightarrow \bar{A}$. Note that this optimal policy is generally *not* implementable in the policy graph, since it may associate two different actions with the same node, depending on the state with which the node is coupled. In other words, we need to know the current state to use this policy. The agent using a policy graph is basically embedded in the cross-product MDP, but it has only partial observability of its product state $(n^t, s^t)$: it sees $n^t$ but not $s^t$. The cross-product MDP is in fact a POMDP $(\bar{S}, \bar{O}, \bar{A}, \bar{B}, \bar{T}, \bar{R})$ where $\bar{O} = N$ and $\bar{B}$ is the projection of $N \times S$ on $N$. However, the solution of the fundamental equation (7) is useful in some algorithms, because it represents an upper-bound of the performance attainable by any implementable policy. We will use this in a branch-and-bound algorithm for finding the optimal deterministic policy graph. Note that the addition of restriction constraints on the policy, as defined in section 2.2, does not invalidate theorem 2. It just limits the set of possible actions in some states of the cross-product MDP, and then it reduces the complexity of its solution. As a consequence, the branch-and-bound algorithm will also be able to find the best graph under some restriction constraints.

**Computational leverage.** It is a very important point that most of the computation performed by the algorithms that will follow consists of solving a Bellman equation with a fixed policy as in (4), or for the sake of finding the optimal deterministic policy as in (7). This can be done by successive approximations, the algorithm being called "value iteration" in the case of (7). The complexity of the algorithm is $O(|\bar{S}|^2 |O| |A|) = O(|N|^2 |S|^2 |O| |A|)$ (times the number of iterations, which can be $O(|\bar{S}|)$). The important point is that any structure in the transition matrix $\bar{T}$ can be exploited while executing these back-ups. The structure of $\bar{T}$ has two components:

the structure of the POMDP: A sparse transition matrix $T$ of the POMDP provides leverage that allows the speed-up of successive-approximation iterations [3]. If $K < |S|$ is the branching factor of the POMDP (i.e., the average number of possible successors of a state) then the complexity can be reduced to $O(|N|^2K|S|)$. For instance, deterministic transitions reduce the complexity by a factor of $|S|$. In the same way, a sparse observation matrix is exploitable. For instance, deterministic observations reduce the complexity by a factor of $|O|$.

the structure of the policy graph: If the leverage gained from the structure of the POMDP is not sufficient, then one can choose to restrict further the search space by imposing structural constraints on the graph, and using this structure to speed up the calculation. An extreme, but often adopted, solution is to look for the best RP. In this case, the gain is a factor of $|O|$ (the complexity is $O(|O|^2|S|^2|A|)$ instead of $O(|N|^2|S|^2|O||A|) = O(|O|^3|S|^2|A|)$). We will say more about constraining the policy graph in section 4.

When both the problem and the policy are structured, the leverage gained can be bigger than just the addition of the effect of the two structures. For instance, evaluating a RP in a completely deterministic problem can be done in $O(|S||A|)$ instead of $O(|N||S||A|) = O(|O||S||A|)$.

# 3  FINDING THE OPTIMAL POLICY GRAPH

In this paper, we consider the problem of finding the best policy graph of a given size for a POMDP. Littman [14] showed that finding the best RP for a given POMDP is an NP-hard problem. First, we generalize this result to any finite policy graph with a given number of nodes and any set of restriction constraints.

**Theorem 3** *Given a* POMDP *and a set of restriction constraints, the problem of finding the optimal deterministic policy graph satisfying the constraints is NP-hard.*

The proof is straightforward: Finding the best deterministic policy graph is equivalent to finding the best deterministic RP of the cross-product POMDP. Then the result follows from Littman's theorem.

The techniques for solving NP-hard problems may be classified into three groups: global search, local search and approximation algorithms. In this paper, we will use two classic techniques, a global search (section 3.1) and a local search (section 3.2). We will consider a possible approximation algorithm in section 3.3.

## 3.1  GLOBAL SEARCH

A heuristically guided search is used to find the best *deterministic* policy graph of a given size, whatever the restriction constraints imposed on $\psi$ (actions) and $\eta$ (structure).

It is a branch-and-bound algorithm; i.e., it systematically enumerates the set of all possible solutions using bounds on the quality of partial solutions to exclude entire regions of the search space. If the lower bound of one partial policy is greater than the upper bounds of others, then it is useless explore these partial policies. Otherwise, each possible extension of them will considered in time. Therefore, the algorithm is guaranteed to find the optimal solution in finite time. Note that this approach is a generalization of a previous algorithm proposed by Littman [14]. His algorithm is limited to policy graphs representing RPs and to POMDPs with a very particular structure: state-transitions and observations are deterministic, and the problem is an achievement task (i.e., there is a given goal state that must be reached as soon as possible). The formalism proposed here handles any kind of POMDP and any kind of policy graph with restriction constraints. However, Littman gives more details on some aspects of the algorithm, and the reader can refer to his paper to complete the brief description that we give here.

**Ordering of free parameters.** The tree of all possible policies is expanded (in depth-first, breadth-first, or in a best-first way) by picking the free parameters of the policy one after the other, and considering all possible assignment values for each of them. The game of pruning some branches based on upper bound/lower bounds comparison is added onto that. The size of the tree that is actually expanded in this process strongly depends on the order in which the free parameters are picked. In our case, it is important that the free values of $\psi$ come before the free values of $\eta$. In other words, when building a solution, we assign actions to the nodes first, and then we fix the structure of the graph. Otherwise, no pruning is possible before all possible structures have been expanded (this is due to the nature of the upper-bound that we use, see below). In our implementation, the parameter $\eta^0$ is expanded after $\psi$ but before $\eta$. There is also an issue with the symmetry of the policy-graph space. For instance, in the absence of restriction constraints, we can permute the role played by the different nodes without changing the policy. Each policy graph is then represented by $|N|!$ leaves of the tree. We can avoid enumerating equivalent graphs by imposing some arbitrary rule when expanding the tree. For instance we can impose that the index of the action attached to a node $i$ always be greater or equal to the index of the action of node $i + 1$, for all $i$. This simple trick can improve greatly the performance of the heuristic search, merely dividing the execution time by $|N|!$.

**Upper bounds.** A partial solution is a general finite policy graph with more restriction constraints than initially (each time we specify an action or a node-transition, we add a constraint). Then we can get an upper bound by solving the cross-product MDP, as explained in the second part of section 2.3, and taking the product $\bar{\pi}^0 \bar{V}^*$. A completely specified policy graph corresponds to an RP of the cross-product (PO)MDP, so no policy graph can do better than the optimal solution of the cross-product MDP. Note that this upper bound has a nice monotonicity property: it does not increase when we fix a free parameter, and it is equal to the true value of the policy graph when the graph is completely specified. On the other hand, as long as no value of $\psi$ is specified, the optimal policy found by solving the cross-product MDP is equivalent to the optimal policy of the original MDP $(S, A, T, R)$: the choice of the action in each $(n, s)$ is independent of $n$ and depends only on $s$. Hence, the calculated upper bound is always equal to the value of the optimal policy of the cross-product MDP. This is why no pruning can be done as long as no value of $\psi$ has been specified and this parameter must be considered first when expending the tree.

**Lower bounds.** If the algorithm searches in depth-first order, then we can use the values of the complete policies already expanded to determine lower bounds on the best performance attainable by extending each partial policy. Otherwise, we can find a lower bound for a given partial policy by completing it at random and calculating the value of the resulting complete policy. An improvement consists of performing a simple local optimization after having completed the policy [14]. In our implementation, we also used a heuristic technique based on the solution of the cross-product MDP to complete the partial policy. We calculate the performance of a complete policy by solving equation (4) in the cross-product MDP.

**Complexity.** The calculation of the upper and lower bounds of each node of the expanded tree requires solving some Bellman equations in the cross-product MDP. Hence it can be done in $O(|N|^2|S|^2)$, or less if there is some structure in the POMDP or the policy. To reduce the number of iterations of successive approximation executed during this calculation, one can store, with each partial policy, the value function found when calculating its upper bound. Then we can start the computation of the upper bound of a child partial policy starting from the value of its parent. Since they are often not very different, we can gain a lot of time with this trick. However, the memory space needed increases dramatically. Even if we can calculate the bounds relatively quickly, the real problem is how many nodes it will be necessary to expand before reaching the optimum. In the worst case, the complete tree of all possible solutions will be expanded, which represents a complexity exponential in the number of degrees of freedom of the policy graph. In practice, our simulations showed that many fewer nodes are actually expanded. Note that adding simple constraints on the policy reduces not only the complexity of the solution of the cross-product MDP, but also the size of the search space and hence the number of nodes expanded.

## 3.2  LOCAL SEARCH

In this approach, we try to find the best *stochastic* policy graph by treating this problem as a classical non-linear numerical optimization problem. Since the value of a policy graph is continuous and differentiable with respect to the policy parameters, we can calculate its gradient and climb it in many different ways. We will not develop all the possibilities for climbing the gradient here, but we will rather focus on the calculation of the gradient, and then just depict a simple implementation of gradient ascent. Note that since the gradient may be calculated exactly, this approach is guaranteed to converge to a local optimum. The topology of the search space, and hence the number of local optima, depends on two things: the structure of the POMDP at hand, and the constraints imposed on the policy. By introducing constraints on the policy, we can hope not only to reduce the execution time of the algorithm, but also to change the "landscape" for a less multimodal one.

**Calculating the gradient.** The value $E^\mu$ of a policy graph $\mu$ is given by equation (6). For each policy parameter $x$ we have $\partial E^\mu / \partial x = \bar{\pi}^0 \partial \bar{V}^\mu / \partial x$. The value function $\bar{V}^\mu$ is given by (5). Hence we have

$$\frac{\partial \bar{V}^\mu}{\partial x} = (I - \gamma \bar{T}^\mu)^{-1} \left[ \frac{\partial \bar{C}^\mu}{\partial x} + \gamma \frac{\partial \bar{T}^\mu}{\partial x} (I - \gamma \bar{T}^\mu)^{-1} \bar{C}^\mu \right].$$

We are interested in the gradient with respect to the policy parameters, i.e., we will consider $x = \psi(n, a)$ and $x = \eta(n, o, n')$. The partial derivative of $\bar{T}^\mu$ and $\bar{C}^\mu$ with respect to these variables can be calculated easily starting from (2) and (3). The main difficulty in the calculation of the gradient is inverting the matrix $(I - \gamma \bar{T}^\mu)$. If we want to exploit the structure of $\bar{T}^\mu$, we can do it by successive approximation, the basic update rule being:

$$W \leftarrow I + \gamma \bar{T}^\mu W, \tag{8}$$

where $W$ is an $|N||S| \times |N||S|$ matrix. Without any useful structure in $\bar{T}^\mu$, the complexity of a complete back up is then in $O(|\bar{S}|^3|O||A|)$. Once the matrix is inverted, the inverse can be used to calculate the gradient with respect to any parameter $x$. A minor acceleration can be obtained by using the value of $(I - \gamma \bar{T}^\mu)^{-1}$ at the previous point to start the iterative computation of this value at the new point. It can reduce the number of iterations of (8) at each step, but it is still a matrix-wise DP with a complexity in $O(|\bar{S}|^3)$, and hence in $O(|S|^3)$.

There is another way of computing the gradient with a complexity only in $O(|\bar{S}|^2)$. Instead of performing the matrix-wise DP to calculate $(I - \gamma \bar{T}^\mu)^{-1}$ explicitly, we perform several (classical) vector-wise DPs for which complexity is in $O(|\bar{S}|^2)$, or less if there is some structure in $N$ or $S$. First we compute $\bar{V}^\mu$ by solving (4), which implies a vector-wise DP with complexity in $O(|\bar{S}|^2|O||A|)$. Then

we calculate

$$\bar{V}_1 \stackrel{\text{def}}{=} \frac{\partial \bar{C}^\mu}{\partial x} + \gamma \frac{\partial \bar{T}^\mu}{\partial x} \bar{V}^\mu.$$

At last, we get $\partial \bar{V}^\mu / \partial x = (I - \gamma \bar{T}^\mu)^{-1} \bar{V}_1$ by iterating

$$\bar{V}_2 \leftarrow \bar{V}_1 + \gamma \bar{T}^\mu \bar{V}_2,$$

which is also a vector-wise, square-complexity DP. The total complexity of this calculation is $O(2|\bar{S}|^2|O||A|)$ (neglecting the calculation of $\bar{V}_2$). Unfortunately, the calculation must be re-done for each policy parameter $x$, since $\bar{V}_1$ and $\bar{V}_2$ depend on $x$. Thus, we have divided the complexity by a factor of $|\bar{S}|$, but multiplied it by the number of free variables of the graph. However, this approach will be useful in most cases, since there are often many fewer free variables in the policy graph than "cross-product states". For instance, if we are looking for the best reactive policy, then the indirect calculation allows us to gain a factor of $|S|/|A|$.

**Climbing the gradient.** Climbing the gradient consists of updating each free value $x$ with the rule $x \leftarrow x + \beta \partial E / \partial x$, where $\beta$ is the step-size parameter. In our case the problem is somewhat more complicated since all the parameters that we optimize are probabilities and we have to ensure that they stay valid (i.e., inside of the simplex) after each update. There are numerous ways for doing that, including renormalizing and using the soft-max function. In our implementation, we chose to project the calculated gradient on the simplex, and then apply it until we reach an edge of the simplex. If we reach an edge and the gradient points outside of the simplex, then we project the gradient on the edge before applying it.

**Related work.** The idea of using a gradient algorithm for solving POMDPs has already been pursued by several authors [2, 12, 27]. The main difference between this work and ours is that these authors use a Monte-Carlo estimation of the gradient instead of an exact calculation, and that they limit themselves to RPs, which is much less general than our approach. Moreover, Jaakkola *et al.* do not use the exponentially discounted criterion (1), but the average reward per time step. In a companion paper [17], we propose a stochastic gradient descent approach for learning finite policy graph during a trial-based interaction with the process.

### 3.3 OTHER APPROACHES

A Monte-Carlo approach based on Watkins'Q-learning [25, 24] is also applicable to our problem. For instance, we can an use Q-learning based on observation-action pairs to find (with no guarantee of convergence) the optimal RP for a POMDP [14]. Another instance is Wiering and Schmidhuber's HQL [26], which learns finite sequences of RPs.

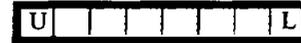

Figure 3: The load/unload problem with 8 locations: the agent starts in the "Unload" location (U) and receives a reward each time it returns to this place after passing through the "Load" location (L). The problem is partially observable because the agent cannot distinguish the different locations in between Load and Unload, and because it cannot perceive if it is loaded or not ($|S| = 14$, $|O| = 3$ and $|A| = 2$).

This Monte-Carlo approach works only if there are strong structural constraints on the graph, and thus cannot be applied for finding general finite policy graphs. Note also that Littman reported observing a great superiority (in terms of execution time) of the global branch-and-bound search over the Monte-Carlo approach, in the case where the graph is constrained to encode a simple RP. Our simulations with other architectures (sets of structural constraints) showed similar results: in general, the Monte-Carlo approach cannot compete with the two others.

## 4   INTRODUCING STRUCTURAL CONSTRAINTS

Because the majority of their computation is to perform Bellman back-ups in the cross-product MDP, the algorithms outlined above can take advantage of any preexisting structure in the POMDP. However, this leverage can be insufficient if the problem is too big or too difficult for the two techniques. In this case, one may whish to restrict further the search space by imposing structural constraints on the policy graph. For instance, a simple solution consists of defining a neighborhood for the nodes of the graph, and allowing transitions only to a neighboring node. This corresponds to a set of restriction constraints ($\eta$ is forced to take the value zero in many points), and hence the algorithm above can still be applied. A somehow extreme solution consists of limiting the search to reactive policies (then $\eta$ is completely fixed in advance). More complex sets of constraints can also be used, for instance, we can limit the search to policy representable as a finite sequence of RPs (with particular rules governing the transition from one RP to another), as in Wiering and Schmidhuber's HQL [26]. Other instances include the finite-horizon memory policies used by McCallum [15, 16], or the external-memory policies used by Peshkin *et al.* [19]. Although these architectures cannot be described only in terms of restriction constraints (there are also equality constraints between different parameters of the graph), the previous results and algorithm can be extended to each of them in particular. In other words, we can use the previous algorithm to find

- the best RP-sequence of a given length,

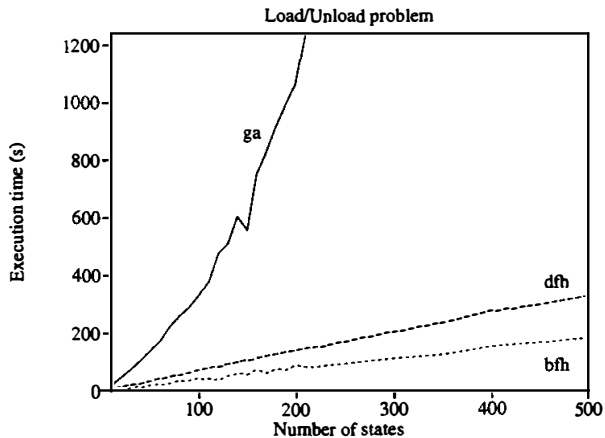

Figure 4: Simulations results obtained with the load/unload problem: execution time of the algorithms as a function of the size of the problem (ga: gradient ascent, dfh: depth first heuristic search, bfh: breadth first heuristic search).

- the best policy using a given finite-horizon memory,

- the best policy using an external-memory of a given size.

(we can also show that it is NP-hard to solve these problems).

What do we gain and what do we lose when we impose a structure on the policy graph? In general, imposing structural constraints reduces the number of parameters per node (which should help both techniques), and modifies the topology of the search space (which influences the gradient descent approach). Another point is that the best graph without the constraints can be better than the best graph with the constraints, i.e., the constraints can decrease the value of the best solution. Even if this does not happen, more nodes may be required to reach the best performance with the constraints than without. Consider, for instance, the load/unload problem represented in figure 3. This simple problem is solved optimally with a two-node policy graph, or with a sequence of two RPs as used in HQL. As an RP is encoded with an $|O|$-node graph, any sequence of two RPs will be encoded by at least $2|O| = 6$ nodes. However, the number of parameters per node will be smaller than in the unconstrained case. In general, adding structure will be interesting if we choose an architecture that fits the problem at hand. Hence, it is a question of previous knowledge about the problem at hand and its optimal solution.

## 5   SIMULATION RESULTS

In our first experiments, we used the simple load/unload problem of figure 3 with an increasing number of locations, to see how both algorithms scale up to increasing problem-size, and how they compare. Since it is a very easy POMDP, the results obtained represent a kind of upper bound on the performance of the algorithms. It is unlikely that they will perform better on another (harder) problem. During this experiment $|N|$ was set to its optimal value of 2 and the gradient algorithm always started from the center of the simplex (i.e., the policy graph is initialized with uniform distributions).[1] We measured the time of execution of each algorithm, as a function of $|S|$. In the case of gradient ascent, we stopped when we reached 99% of the optimal. When the heuristic search uses a stochastic calculation of upper bounds, we average the measure over 50 runs. $\gamma$ was set to 0.996 (a big value is necessary to accommodate big state-spaces), and the learning rate of gradient descent was optimized. The results are given in figure 4. They show that the heuristic search clearly outperforms the gradient algorithm, which becomes numerically unstable when the number of states increases in this kind of geometrically-discounted absorbing problem.

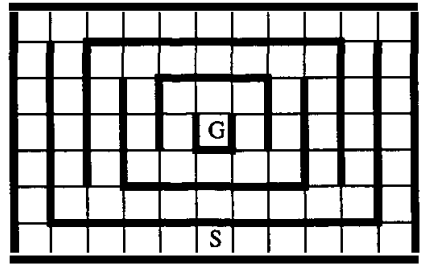

Figure 5: A partially observable stochastic maze: the agent must go from the starting state marked with an "S"to the goal marked with an "G". The problem is partially observable because the agent cannot perceive its true location, but only its orientation and the presence or the absence of a wall on each side of the square defining its current state. The problem is stochastic because there is a non-zero probability of slipping, so that the agent does not always know if its last attempt to make a move had any consequence on its actual position in the maze.

In the second set of experiments, we wanted to measure how far our algorithms can go in terms of problem size, in a problem more difficult than the simple load/unload. We used the a set of partially observable mazes with the regular structure represented in figure 5, and whose size varies from 9 to 989 states ($|O| = 9$ and $|A| = 4$). These mazes are not particularly easy, since they have only two different optimal paths. The minimal number of nodes for solving them is 4, one per action (although the policy is not reactive). The time required for the (depth first) branch-and-bound algorithm to find the optimal solution with this optimal number of nodes is shown in figure 6. We see that we can solve a partially observable maze with almost 1000 states in a less than 6 hours. It represents a performance far above the capacities of classic approaches for solving POMDPs. Note also that, as the number of states grows, the measured complexity is almost linear in the number of states.

---

[1]We used a simpler version of the algorithms where the starting node is fixed. Otherwise, the policy using only uniform distributions is a (very unstable) local optimum.

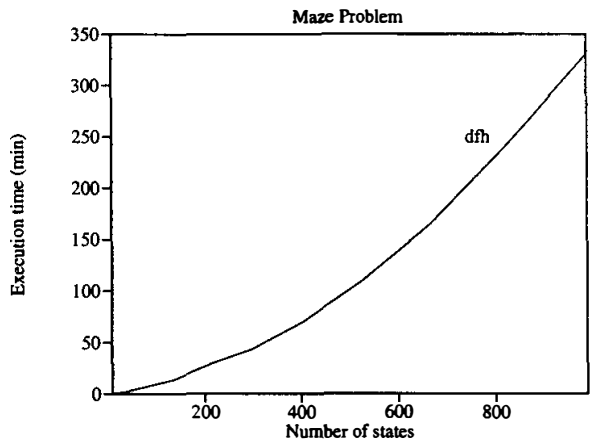

Figure 6: Performance of the branch-and-bound algorithm on the maze problem: execution time as a function of the number of states.

## 6   CONCLUSION

We studied the problem of finding the optimal policy representable as a finite state automaton of a given size, possibly with some simple structural constraints. This approach by-passes the continuous and intractable belief-state space. However, we showed that we end up with a NP-hard problem anyway. Then we proposed to use two classic search techniques, and developed efficient implementations of them that allow using the structure of the problem to accelerate the computation. If this is not sufficient, bigger leverage can be gained by imposing structure on the policy. However, our algorithms are limited by necessity to enumerate at least once per iteration, the complete state space of the POMDP. In a companion paper [17], we propose an indirect learning algorithm that avoids this bottleneck.

## References

[1] K.J. Astrom. Optimal control of Markov decision processes with incomplete state estimation. *J. Math. Anl. Appl.*, 10, 1965.

[2] L.C. Baird and A.W. Moore. Gradient descent for general reinforcement learning. In *Advances in Neural Information Processing Systems, 12*. MIT Press, Cambridge, MA, 1999.

[3] C. Boutillier, T.L. Dean, and S. Hanks. Decision theoretic planning: structural assumptions and computa-

tional leverage. *Journal of AI Research*, To appear, 1999.

[4] A.R. Cassandra. *Exact and Approximate Algorithms for Partially Observable Markov Decision Processes*. PhD thesis, Brown University, 1998.

[5] A.R. Cassandra, L.P. Kaelbling, and M.L. Littman. Acting optimally in partially observable stochastic domains. In *Proceedings of the Twelfth National Conference on Artificial Intelligence*, 1994.

[6] E.A. Hansen. An improved policy iteration algorithm for partially observable MDPs. In *Advances in Neural Information Processing Systems, 10*. MIT Press, Cambridge, MA, 1997.

[7] E.A. Hansen. *Finite-Memory Control of Partially Observable Systems*. PhD thesis, Department of Computer Science, University of Massachusetts at Amherst, 1998.

[8] E.A. Hansen. Solving POMDPs by searching in policy space. In *Proceedings of the Eighth Conference on Uncertainty in Artificial Intelligence*, pages 211–219, Madison, WI, 1998.

[9] M. Hauskrecht. *Planning and Control in Stochastic Domains with Imperfect Information*. PhD thesis, MIT, Cambridge, MA, 1997.

[10] O. Higelin. *Optimal Control of Complex Structured Processes*. PhD thesis, University of Caen, France, 1999.

[11] R.A. Howard. *Dynamic Programming and Markov Processes*. MIT Press, Cambridge, 1960.

[12] T. Jaakkola, S. Singh, and M.R. Jordan. Reinforcement learning algorithm for partially observable Markov problems. In *Advances in Neural Information Processing Systems, 7*. MIT Press, Cambridge, MA, 1994.

[13] L.P. Kaelbling, M.L. Littman, and A.R. Cassandra. Planning and acting in partially observable stochastic domains. *Artificial Intelligence*, 101, 1998.

[14] M.L. Littman. Memoryless policies: Theoretical limitations and practical results. In *From Animals to Animats 3: Proceedings of the Third International Conference on Simulation of Adaptive Behavior*. MIT Press, Cambridge, MA, 1994.

[15] R.A. McCallum. Overcoming incomplete perception with utile distinction memory. In *The Proceedings of the Tenth International Machine Learning Conference*, Amherst, MA, 1993.

[16] R.A. McCallum. *Reinforcement Learning with Selective Perception and Hidden State*. PhD thesis, University of Rochester, Rochester, NY, 1995.

[17] N. Meuleau, L. Peshkin, K.E. Kim, and L.P. Kaelbling. Learning finite-state controllers for partially observable environments. *Proceedings of the Fifteenth Conference on Uncertainty in Artificial Intelligence*, To appear, 1999.

[18] R. Parr and S. Russell. Reinforcement learning with hierarchies of machines. In *Advances in Neural Information Processing Systems 11*. MIT Press, Cambridge, MA, 1998.

[19] L. Peshkin, N. Meuleau, and L.P. Kaelbling. Learning policies with external memory. *Proceedings of the Sixteenth International Conference on Machine Learning*, To appear, 1999.

[20] M.L. Puterman. *Markov Decision Processes: Discrete Stochastic Dynamic Programming*. Wiley, New York, NY, 1994.

[21] J.K. Satia and R.E. Lave. Markov decision processes with probabilistic observation of states. *Management Science*, 20(1):1–13, 1973.

[22] R.D. Smallwood and E.J. Sondik. The optimal control of partially observable Markov decision processes over a finite horizon. *Operations Research*, 21:1071–1098, 1973.

[23] E.J. Sondik. The optimal control of partially observable Markov decision processes over the infinite horizon: Discounted costs. *Operations Research*, 26, 1978.

[24] R.S. Sutton and A.G. Barto. *Reinforcement Learning: An Introduction*. MIT Press, Cambridge, MA, 1998.

[25] C. Watkins. *Learning from Delayed Rewards*. PhD thesis, King's College, Cambridge, 1989.

[26] M. Wiering and J. Schmidhuber. HQ-Learning. *Adaptive Behavior*, 6(2):219–246, 1997.

[27] R.J. Williams. Towards a theory of reinforcement-learning connectionist systems. Technical Report NU-CCS-88-3, Northeastern University, Boston, MA, 1988.